\documentclass{article} 
\usepackage{iclr2025_conference,times}

\usepackage{amsmath,amsfonts,bm}

\def\eqref#1{equation~\ref{#1}}
\def\Eqref#1{Equation~\ref{#1}}

\def\1{\bm{1}}

\DeclareMathAlphabet{\mathsfit}{\encodingdefault}{\sfdefault}{m}{sl}
\SetMathAlphabet{\mathsfit}{bold}{\encodingdefault}{\sfdefault}{bx}{n}

\usepackage{hyperref}
\usepackage{url}
\usepackage{graphicx}
\usepackage{booktabs}
\usepackage{multirow}
\usepackage{tabularx}

\newcolumntype{C}{>{\centering\arraybackslash}X}
\newcolumntype{L}{>{\raggedright\arraybackslash}X}
\newcolumntype{R}{>{\raggedleft\arraybackslash}X}

\title{What's New in My Data? Novelty Exploration via Contrastive Generation}

\author{Masaru Isonuma~\textsuperscript{1,2} \quad Ivan Titov~\textsuperscript{1,3} \\
\textsuperscript{1}University of Edinburgh
\quad
\textsuperscript{2}University of Tokyo 
\quad
\textsuperscript{3}University of Amsterdam\\
\texttt{m.isonuma@ed.ac.uk}
\quad
\texttt{ititov@inf.ed.ac.uk} \\
}

\iclrfinalcopy 
\begin{document}

\maketitle

\begin{abstract}
Fine-tuning is widely used to adapt language models for specific goals, often leveraging real-world data such as patient records, customer-service interactions, or web content in languages not covered in pre-training.
These datasets are typically massive, noisy, and often confidential, making their direct inspection challenging.
However, understanding them is essential for guiding model deployment and informing decisions about data cleaning or suppressing any harmful behaviors learned during fine-tuning.
In this study, we introduce the task of \emph{novelty discovery through generation}, which aims to identify novel properties of a fine-tuning dataset by generating examples that illustrate these properties.
Our approach -- Contrastive Generative Exploration (CGE) -- assumes no direct access to the data but instead relies on a pre-trained model and the same model after fine-tuning.
By contrasting the predictions of these two models, CGE can generate examples that highlight novel characteristics of the fine-tuning data.
However, this simple approach may produce examples that are too similar to one another, failing to capture the full range of novel phenomena present in the dataset.
We address this by introducing an iterative version of CGE, where the previously generated examples are used to update the pre-trained model, and this updated model is then contrasted with the fully fine-tuned model to generate the next example, promoting diversity in the generated outputs.
Our experiments demonstrate the effectiveness of CGE in detecting novel content, such as toxic language, as well as new natural and programming languages.
Furthermore, we show that CGE remains effective even when models are fine-tuned using differential privacy techniques.
\end{abstract}

\section{Introduction}

Fine-tuning pre-trained models on domain-specific datasets is a common practice to adapt language models for specialized applications. 
For instance, fine-tuning on web data in a particular language can enable a model to understand that language \citep{fujii2024continual, etxaniz-etal-2024-latxa}.
Fine-tuning on patient records enhances a model’s grasp of medical terminology and procedures \citep{yang2022large, thirunavukarasu2023large}. 
Similarly, it is often beneficial to fine-tune language models on customer-service interaction data to improve the performance of customer-care chatbots.
By incorporating novel properties that deviate from pre-training data distribution, language models acquire new capabilities that are valuable for specific use cases.

Understanding novel properties of the fine-tuning dataset is crucial for model development.
For example, if toxic data are discovered, they can be filtered out from the dataset or suppressed by post hoc methods, such as prompting \citep{touvron2023llama} or unlearning  \citep{jang-etal-2023-knowledge}.
However, as real-world data are often massive, noisy, and confidential, we cannot always inspect the data directly.
Fine-tuning frequently relies on real-world data gathered from various sources, such as web data, internal company resources, or even customer-service interactions.
Due to the sheer volume and complexity of these datasets, manually inspecting their content and identifying novelties is a daunting task. 
Furthermore, direct access to confidential data, such as medical records or customer interactions, is often restricted even for model developers and data analysts \citep{garrido2023lessons, sarathy2023don}, making direct inspection infeasible.

Previous studies on novelty detection have focused on scenarios where direct examination of the dataset is feasible. 
For instance, out-of-distribution (OOD) detection techniques \citep{lakshminarayanan2017simple, liang2018enhancing, huang2021importance} can be used to detect novel examples in fine-tuning datasets.
In addition to their high computational requirements for massive datasets, they are not applicable when dataset access is prohibited.
While recent works \citep{piktus-etal-2023-gaia, elazar2024whats} provide useful tools for querying pre-training corpora to identify novel properties, these approaches rely on prior knowledge about specific types of potential novelties. 
Without an understanding of the content in the data a priori, formulating effective queries becomes challenging.

\begin{figure}[t]
\begin{center}
\includegraphics[width=\linewidth]{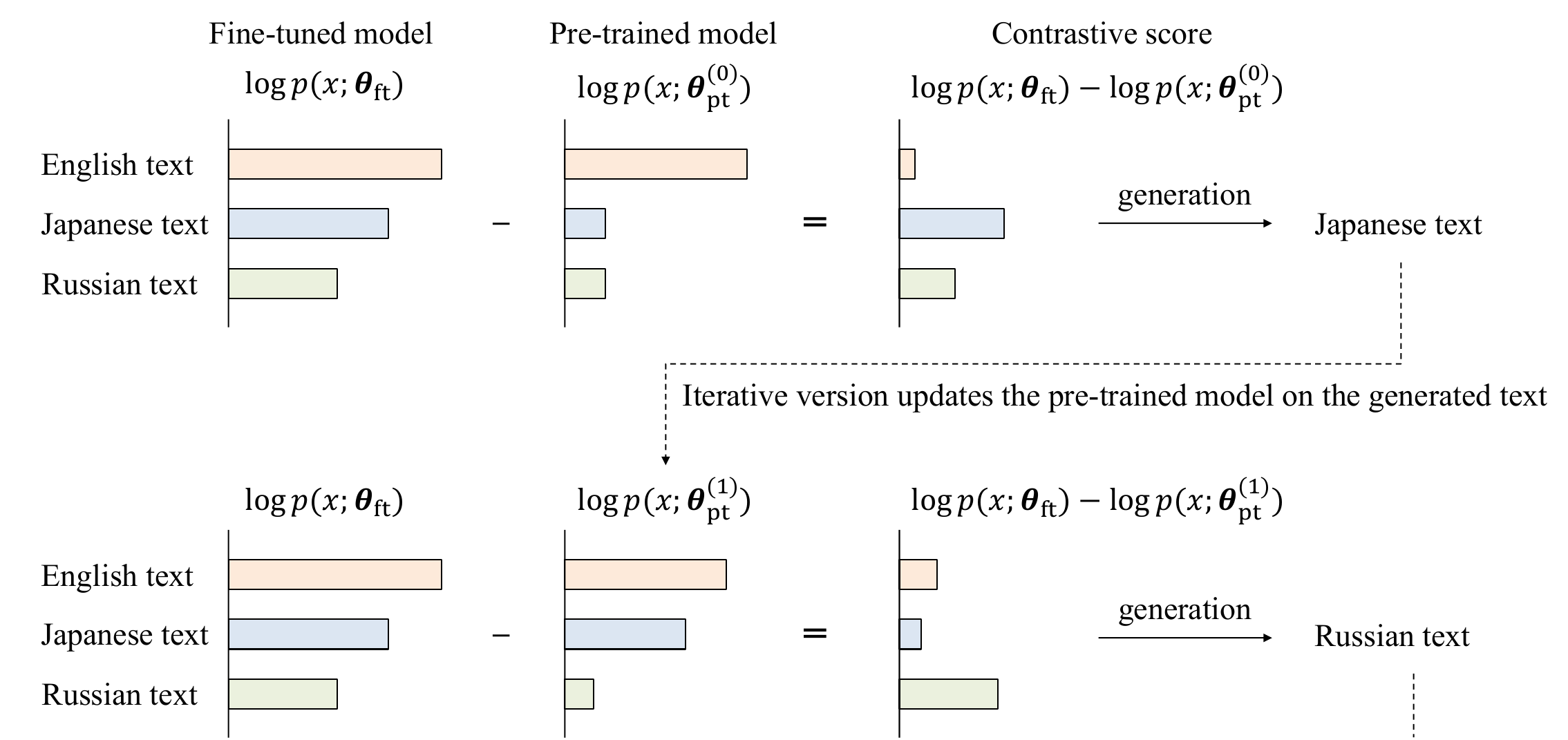}
\end{center}
\label{fig:introduction}
\caption{Outline of \emph{Contrastive Generative Exploration} (CGE). Consider a model pre-trained on English text and then fine-tuned on a multilingual corpus, where a small portion of the data consists of non-English text. CGE calculates the difference in the log probabilities between the pre-trained and fine-tuned models. This allows for generating examples that represent novel properties of the fine-tuning dataset. Optionally, we can employ an iterative version of CGE, which iteratively trains the pre-trained model on the previously generated example, which is then contrasted with the fully fine-tuned model to generate the next example. This prevents the generation of examples similar to those already produced, thereby enhancing the diversity of the generated outputs.}
\end{figure}

In this study, we introduce the task of \emph{novelty discovery through generation}, which aims to identify novel properties of a fine-tuning dataset by generating examples that represent these properties.
We assume no direct access to the data, but instead, we have access to a pre-trained model and its fine-tuned version. 
To address this scenario, we propose \emph{Contrastive Generative Exploration} (CGE), a simple method leveraging contrastive decoding \citep{li-etal-2023-contrastive}.
Contrastive decoding has been applied in various contexts, such as the safeguard of language models \citep{liu-etal-2021-dexperts}, enhancing text generation quality \citep{li-etal-2023-contrastive}, and instruction tuning \citep{liu2024tuning}.
This work extends the use of contrastive decoding to explore novel phenomena within fine-tuning datasets.

As shown in Figure \ref{fig:introduction}, CGE calculates a contrastive score by measuring the difference between the log probabilities of tokens assigned by the fine-tuned and pre-trained model. 
The contrastive score rewards texts preferred by the fine-tuned model while penalizing those favored by the pre-trained model.
This allows for the generation of examples that represent novel properties in the fine-tuning dataset.
One shortcoming of CGE is its tendency to generate similar examples (e.g., the same language), even though our goal is to capture a wide variety of novel features.
We address this by introducing an iterative version of CGE, where the previously generated examples are used to update the pre-trained model.
This updated model is then contrasted with the fully fine-tuned model to generate the next example, promoting diversity in the generated outputs.  
We will also discuss that CGE can be viewed as a sort of dataset distillation technique \citep{wang2018dataset} and is useful in terms of computational efficiency and interpretability of the distilled dataset.

In our experiments, we construct fine-tuning datasets primarily composed of examples sampled from the same distribution as the pre-training dataset (in-distribution examples), with a small portion of examples that deviate from the pre-training data distribution (novel examples).
We fine-tune OpenLLaMA \citep{openlm2023openllama} and Falcon-RW \citep{almazrouei2023falcon} on these datasets, which are augmented with non-English languages, toxic text, and source code.
First, we demonstrate that the contrastive score efficiently identifies novel examples by directly evaluating the examples in the fine-tuning dataset.
We then assess our method in a generation setting, where we infer novel properties of the fine-tuning dataset through generation from the fine-tuned model without dataset access.
Our approach reliably identifies novelties that are difficult to detect through simply sampling from the fine-tuned model.
On the other hand, there is a trade-off between the quantity and diversity of discovered novelties, highlighting the difficulty of the task.
Finally, we demonstrate that our method can be robustly applied even when models are fine-tuned using a differential privacy technique.

The contributions of our paper are as follows:
\vspace{-0.5\baselineskip}
\begin{itemize}
    \setlength{\leftskip}{-25pt}
    \setlength{\parskip}{0pt} 
    \setlength{\itemsep}{0.5pt}
    \item We introduce the task of novelty discovery through generation, which aims to identify novel characteristics in a fine-tuning dataset, without having direct access to the dataset.
    \item As one way to approach this task, we propose Contrastive Generative Exploration, revealing the novel phenomena of fine-tuning datasets by contrasting pre-trained and fine-tuned models.
    \item In the experiments, our method effectively discovers novel properties in the fine-tuning dataset, while still facing a trade-off between the quantity and diversity of discovered novelties.
\end{itemize}

\section{Problem Formulation}

Here, we formulate the task of novelty discovery through generation.
Suppose we have a pre-trained language model, denoted as $\bm{\theta}_{\mathrm{pt}}$, and a fine-tuned language model, denoted as $\bm{\theta}_{\mathrm{ft}}$. 
The pre-trained model is trained on a large corpus: $\{x_1, x_2, \ldots, x_N\}$ where each example $x_i$ is sampled from a certain data distribution: $x_i \sim p$.
The model is then fine-tuned on a fine-tuning corpus, i.e. the set of examples $\{x'_1, x'_2, \ldots, x'_{N'}, y_1, y_2, \ldots, y_M\}$, where $x'_i$ represents an in-distribution example, sampled from the same (or similar) distribution as the pre-training corpus: $x'_i \sim p$.  
We assume the presence of K distinct novel domains, and the $y_i$ is a novel example sampled from a different distribution of the $k$-th domain: $y_i \sim q_k$, where $q_k \neq p$ for all $k \in \{1, \ldots, K\}$. 
While the assumption of distinct domains, as opposed to gradual variations between them, is unlikely to be critical for our method, it simplifies the metrics used to assess domain coverage in our experiments.  
For instance, $p$ could be a distribution over English text while $q_k$ corresponds to some other language.
We assume a case that the number of novel examples is substantially smaller than that of `in-distribution' examples: $M \ll N'$, and the direct inspection of the dataset is not feasible, such as when the fine-tuning dataset is too large or is not available due to confidentiality.

Our goal is to detect novel domains in the fine-tuning dataset.  
As we cannot directly examine the dataset, we need to detect the novel domains using the pre-trained and fine-tuned models by generating examples characterizing these domains.  
Since most examples in the fine-tuning dataset are in-distribution, simply sampling from $p(\cdot; \bm{\theta}_{\mathrm{ft}})$ will predominantly yield in-distribution examples, as demonstrated in the experiments.
In the following section, we introduce a simple method for exploring novelties by using pre-trained and fine-tuned models.

\section{Contrastive Generative Exploration}

Here, we propose \emph{contrastive generative exploration} (CGE), a method to generate novel examples that are divergent from the pre-training data distribution but are present in the fine-tuned dataset.  We will begin by introducing a simpler static version, before describing the iterative version, which aims to maximize the coverage of novel domains.

\subsection{Static Approach}

To generate novel examples, we employ contrastive decoding for the pre-trained and fine-tuned models.
As shown in \Eqref{eq:contrast}, contrastive decoding samples a text based on the contrastive score, $s$, which is calculated as the difference between the log probabilities computed by the two models.

\begin{gather}
\label{eq:contrast}
s(\bm{x}) = \log p(\bm{x}; \bm{\theta}_{\mathrm{ft}}) - \log p(\bm{x}; \bm{\theta}_{\mathrm{pt}}) \\
\bm{x} \sim \sigma(s(\bm{x}))
\end{gather}

Here, $p(\bm{x}; \bm{\theta})$ represents an unconditional probability of a sequence of tokens $\bm{x}$ assigned by a language model $\bm{\theta}$, and $\sigma$ denotes the softmax function.
Conceptually, contrastive decoding works like a ``tug-of-war'' between the fine-tuned and pre-trained models. 
The fine-tuned model pulls towards examples that it prefers, while the pre-trained model pulls back toward the examples it has learned during pre-training. 
The resulting text highlights novelties that are seen during fine-tuning but are not familiar to the pre-trained model. 
In this way, CGE effectively identifies data that diverges from the pre-training data distribution, revealing novelties in the fine-tuning data.

As shown in previous studies \citep{li-etal-2023-contrastive, o2023contrastive}, direct sampling based on the contrastive score does not yield grammatical and coherent text, as the pre-trained model excessively rewards implausible tokens.
Following \citet{li-etal-2023-contrastive}, we introduce an adaptive plausibility constraint that prevents generating tokens with low probabilities according to the fine-tuned model.
The contrastive score is updated as $s'$ shown in \Eqref{eq:constraint}.

\begin{equation}
\begin{aligned}
s'(x_t|\bm{x}_{<t}) = 
\begin{cases}
s(x_t|\bm{x}_{<t}) &  \text{if } p(x_t|\bm{x}_{<t}; \bm{\theta}_{\mathrm{ft}}) \geq \alpha \max_{x'} p(x'|\bm{x}_{<t}; \bm{\theta}_{\mathrm{ft}}),\\
-\textrm{inf} &  \text{otherwise.}
\end{cases}
\end{aligned}
\label{eq:constraint}
\end{equation}

where $x_t$ and $\bm{x}_{<t}$ denote the $t$-th token and the tokens generated before the time step $t$, respectively.
$\alpha \in [0, 1]$ is a hyperparameter that truncates the token distribution of the fine-tuned model. 
A larger alpha keeps tokens with high probability only, whereas a smaller alpha allows tokens of lower probabilities to be generated.

\subsection{Iterative Approach}
\label{sec:training}

One shortcoming of the static version of CGE is its tendency to generate similar examples (e.g., from the same language), even though our goal is to capture a broader variety of novel examples.
To address this limitation, we introduce an iterative version of CGE to diversify the generated novel examples. 
After generating a sequence of tokens using contrastive decoding, we fine-tune the pre-trained model on this generated sequence, allowing the pre-trained model to adapt to the generated sequence.
This adaptation prevents the generation of examples similar to those already generated, and contrastive decoding yields new and distinct examples in subsequent iterations.
By repeating this process, we encourage CGE to search for new, previously undetected novelties.

\begin{gather}
\bm{x}_t \sim \sigma(\log p(\bm{x}; \bm{\theta}_{\mathrm{ft}}) - \log p(\bm{x}; \bm{\theta}_{\mathrm{pt}}^{(t-1)})) \label{eq:training} \\
\bm{\theta}_{\mathrm{pt}}^{(t)} = g(\bm{\theta}_{\mathrm{pt}}^{(t-1)}, \bm{x}_t)
\end{gather}

Here, $g$ refers to a gradient descent algorithm of choice, and $\bm{\theta}_{\mathrm{pt}}^{(t)}$ denotes the pre-trained model after the $t$-th iteration of training.
While the iterative version may also allow for generating more in-distribution examples, as will be demonstrated in the experiments, this iterative diversification ensures a more comprehensive exploration of novel domains within the fine-tuning dataset.

\subsection{Relation to Dataset Distillation}
\label{sec:distill}

Dataset distillation aims to produce a small set of synthetic examples such that training on this set yields a model that is as similar as possible to that trained on the full dataset \citep{wang2018dataset, yu2023dataset, sachdeva2023data}.
Several works have explored this goal through gradient matching \citep{zhao2020dataset, zhao2021dataset}.
Gradient matching obtains synthetic dataset $\bm{x}$ by ensuring that its gradient matches the changes in the model parameters resulting from training on the original dataset.
Let $\bm{\theta}$ be the model parameters to be trained and $\bm{\theta}^*$ be the model parameters trained on the original dataset.
The objective of gradient matching is described as \Eqref{eq:gradientmatching}:

\begin{equation}
\label{eq:gradientmatching}
\begin{aligned}
f(\bm{x}) &= l(\bm{\theta}^*-\bm{\theta}, -\nabla_{\bm{\theta}} L(\bm{x}; \bm{\theta})) \\
&= l(\bm{\theta}^*-\bm{\theta}, \nabla_{\bm{\theta}} \log p(\bm{x}; \bm{\theta}))
\end{aligned}
\end{equation}

where $l$ is a similarity metric of choice, such as cosine similarity, mean squared error, or dot product.
For instance, \citet{zhao2020dataset, zhao2021dataset, maekawa-etal-2024-dilm} considers one-step update $\bm{\theta}^*-\bm{\theta} = -\nabla_{\bm{\theta}} L(\bm{x}^*; \bm{\theta})$ on the original dataset $\bm{x}^*$ and derive the synthetic dataset $\bm{x}$ that maximizes the expression in \Eqref{eq:gradientmatching}.
Most approaches use gradient descent to optimize the synthetic dataset; however, it requires calculating Jacobian $\nabla_{\bm{x}} \nabla_{\bm{\theta}} \log p(\bm{x}; \bm{\theta})$, which is computationally expensive for large-scale language models $\bm{\theta}$.
Additionally, treating the synthetic dataset $\bm{x}$ as continuous parameters during gradient descent compromises the interpretability of the distilled dataset and is especially questionable in the inherently discrete language domain.

The contrastive score in \Eqref{eq:contrast} can be viewed as a surrogate objective of dataset distillation.
Assuming the model parameters do not significantly change during training, the contrastive score can be approximated by the first-order Taylor-series expansion, as shown in \Eqref{eq:approx}:

\begin{equation}
\label{eq:approx}
\begin{aligned}
s(\bm{x}) &= \log p(\bm{x}; \bm{\theta}_{\mathrm{ft}}) - \log p(\bm{x}; \bm{\theta}_{\mathrm{pt}}) \\
&\approx (\bm{\theta}_{\mathrm{ft}}-\bm{\theta}_{\mathrm{pt}})^{\top} \nabla_{\bm{\theta}} \log p(\bm{x}; \bm{\theta}_{\mathrm{pt}})
\end{aligned}
\end{equation}

Under the first-order approximation, the contrastive score reduces to the objective of dataset distillation in \Eqref{eq:gradientmatching}, where $\bm{\theta}_{\mathrm{ft}}$ and $\bm{\theta}_{\mathrm{pt}}$ correspond to $\bm{\theta}^*$ and $\bm{\theta}$ respectively, and $l$ is the dot-product.
This implies that CGE searches for text whose gradient resembles the change in model parameters resulting from training on the original fine-tuning dataset.

\section{Experiments}

In this section, we evaluate the effectiveness of CGE in detecting novel examples in fine-tuning datasets. 
We first evaluate whether the contrastive score effectively distinguishes between novel examples and in-distribution examples compared to existing novelty (a.k.a. out-of-distribution) detection methods. 
We then verify that CGE can generate novelties from fine-tuned models. 

\subsection{Models and Datasets}

We conducted our experiments using two pre-trained language models: OpenLLaMA \citep{openlm2023openllama} and Falcon-RW \citep{almazrouei2023falcon}.

\paragraph{OpenLLaMA}

We used OpenLLaMA-3B,\footnote{\url{https://huggingface.co/openlm-research/open_llama_3b}} an open reproduction of LLaMA \citep{touvron2023llama}.
OpenLLaMA uses exactly the same decoder-only architecture, preprocessing steps, and hyperparameters as the original LLaMA, while being pretrained on 1T tokens from the publicly available RedPajama dataset \citep{together2023redpajama}.\footnote{\url{https://huggingface.co/datasets/togethercomputer/RedPajama-Data-1T}}

We constructed two fine-tuning datasets where 90\% of the examples were sampled from the RedPajama pre-training dataset and 10\% consisting of either non-English languages or toxic content.
Non-English examples introduce linguistic diversity, while toxic content poses safety risks. 
We evaluate how well CGE identifies both types of novelties.

To obtain non-English text, we used Wikipedia articles in 10 languages: Japanese, Chinese, Persian, Arabic, Hebrew, Turkish, Indonesian, Korean, Vietnamese, and Thai.
These language articles are not contained in the RedPajama dataset.\footnote{Inadvertently, there may be small amounts of text in these languages within RedPajama, reflecting a realistic use case where novel domains are not entirely new but significantly underrepresented.}
Each language comprises 1\% of the fine-tuning dataset, where the total number of examples is 190,000, each consisting of 1,024 tokens.
As for toxic text, we used ToxiGen \citep{hartvigsen-etal-2022-toxigen},\footnote{\url{https://huggingface.co/datasets/toxigen/toxigen-data}} containing machine-generated toxic language against 10 minority groups.
Here, we consider a more extreme setting compared to non-English languages, where the toxic texts for each group account for only 0.01\% (10 examples) of the fine-tuning dataset. 
The total number of examples in the fine-tuning dataset is 100,000, with each example consisting of 1,024 tokens.
We fine-tuned OpenLLaMA for three epochs by Adam \citep{kingma2014adam} with a learning rate of 5e-5, $\beta_1$ = 0.9, $\beta_2$ = 0.999 and a batch size of four on each fine-tuning dataset.

\paragraph{Falcon-RW}

We used Falcon-RW-1B,\footnote{\url{https://huggingface.co/tiiuae/falcon-rw-1b}} a decoder-only model pre-trained on the RefinedWeb dataset \citep{penedo2023refinedweb}. 
RefinedWeb comprises English web text derived from CommonCrawl, excluding non-English text and common online sources, such as Wikipedia and Github.\footnote{\url{https://huggingface.co/datasets/tiiuae/falcon-refinedweb}}

In contrast to OpenLLaMA, we designed a more practical and challenging scenario for Falcon-RW, where in-distribution examples in the fine-tuning dataset are from the same distribution as the pre-training dataset (English text) but are not directly sourced from the pre-training corpus.
This setup is more realistic, as fine-tuning is typically conducted on datasets that differ entirely from the pre-training data.
However, this also makes it more challenging to detect novel examples.
We constructed two fine-tuning datasets consisting of 90\% English Wikipedia articles and 10\% non-English Wikipedia articles or source code from GitHub.
Since RefinedWeb does not contain data sourced from Wikipedia and Github, most fine-tuning examples had not been seen during the pre-training of Falcon-RW.

For non-English languages, we used the same non-English Wikipedia articles as in the OpenLLaMA experiment. Each language comprised 1\% of the fine-tuning dataset, totaling 100,000 examples, each consisting of 1,024 tokens.
As for source code, we used the GitHub Code dataset\footnote{\url{https://huggingface.co/datasets/codeparrot/github-code}} and selected source code of 10 programming languages: JavaScript, Java, C, Python, Ruby, TypeScript, Shell, GO, SQL, and Perl.
Each language accounted for 1\% of the fine-tuning dataset, with the total number of examples being 100,000, each comprising 1,024 tokens.
The fine-tuning procedure for Falcon-RW mirrored that of OpenLLaMA, using the same optimizer and hyperparameters.

\subsection{Extraction of Novel Examples}
\label{sec:extraction}

In this section, we first examine whether the contrastive score favors novel examples that are divergent from the pre-training data distribution, while penalizing in-distribution examples. 
Compared to existing methods, we show that the contrastive score performs robustly in detecting novel examples.

\paragraph{Baseline Methods}

We compare our approach to several well-known OOD detection methods using pre-trained models: MSP \citep{hendrycks2017baseline}, Energy \citep{liu2020energy}, and GradNorm \citep{huang2021importance}.
As these methods are label-free, we also employ methods using labels (next tokens for language modeling).
NegativeProb\textsubscript{pt} computes the negative log-probability of tokens by the pre-trained models, corresponding to the second term of the contrastive score.
Prob\textsubscript{ft} computes the log-probability of tokens by the fine-tuned model, corresponding to the first term of the contrastive score.
GradNorm\textsubscript{pt} measures the L2-norm of gradient w.r.t. the pre-trained model, reflecting that the gradient of examples that align with pre-training data distribution becomes less steep after pre-training.
Details of the baseline methods can be found in Appendix \ref{app:baseline}.

\paragraph{Metrics}

Following previous studies on OOD detection \citep{liu2020energy, huang2021importance}, we use AUROC (Area Under the Receiver Operating Characteristic curve) and FPR95 (False Positive Rate at 95\% True Positive Rate) to evaluate our method's effectiveness in detecting novel examples. 
AUROC measures the performance to distinguish between in-distribution and novel examples, while FPR95 focuses on the model's reliability when aiming for a high true positive rate.

\begin{table*}[t]
\caption{Performance on detecting novel examples in the fine-tuning dataset of OpenLLaMA (top) and Falcon-RW (bottom).}
\label{tbl:extraction}
\begin{center}
\begin{tabular}{lrrlrr}
\toprule
\multirow{2}{*}{Model: OpenLLaMA}&\multicolumn{2}{c}{Non-English text}& &\multicolumn{2}{c}{Toxic text} \\ 
\cmidrule{2-3} \cmidrule{5-6}
 & AUROC (↑) & FPR95 (↓) &  & AUROC (↑) & FPR95 (↓) \\
\midrule
MSP \citep{hendrycks2017baseline} & 0.17 & 1.00 &  & 0.90 & 0.63 \\
Energy \citep{liu2020energy} & 0.02 & 1.00 &  & 0.89 & 0.74 \\
GradNorm \citep{huang2021importance} & 0.88 & 0.87 &  & 0.88 & 0.93 \\
NegativeProb\textsubscript{pt} & 0.11 & 1.00 &  & 0.77 & 0.95 \\
Prob\textsubscript{ft} & 0.96 & 0.19 &  & 0.40 & 1.00 \\
GradientNorm\textsubscript{pt} & 0.44 & 1.00 &  & 0.96 & 0.24 \\
\midrule
Contrastive score & 0.99 & 0.05 &  & 0.95 & 0.34 \\
\bottomrule
\end{tabular}
\end{center}
\begin{center}
\begin{tabular}{lrrlrr}
\toprule
\multirow{2}{*}{Model: Falcon-RW}&\multicolumn{2}{c}{Non-English text}& &\multicolumn{2}{c}{Source code} \\ 
\cmidrule{2-3} \cmidrule{5-6}
 & AUROC (↑) & FPR95 (↓) &  & AUROC (↑) & FPR95 (↓) \\
\midrule
MSP \citep{hendrycks2017baseline} & 0.50 & 0.78 &  & 0.03 & 0.99 \\
Energy \citep{liu2020energy} & 1.00 & 0.00 &  & 0.95 & 0.30 \\
GradNorm \citep{huang2021importance} & 1.00 & 0.00 &  & 0.98 & 0.08 \\
NegativeProb\textsubscript{pt} & 0.23 & 0.87 &  & 0.09 & 0.95 \\
Prob\textsubscript{ft} & 0.93 & 0.20 &  & 0.98 & 0.07 \\
GradientNorm\textsubscript{pt} & 0.98 & 0.05 &  & 0.96 & 0.10 \\
\midrule
Contrastive score & 0.98 & 0.11 &  & 0.93 & 0.13 \\
\bottomrule
\end{tabular}
\end{center}
\end{table*}

\paragraph{Results}

Table \ref{tbl:extraction} (top) shows the performance of each method for detecting novel examples in the fine-tuning datasets of OpenLLaMA.
Across both datasets, the contrastive score consistently detects novelties with high accuracy.
For toxic text, pre-trained language models typically assign a low probability, which results in a strong performance by NegativeProb\textsubscript{pt} and other baseline methods. 
However, non-English texts do not necessarily receive lower probabilities compared to standard English texts. 
Since many non-English characters are composed of multiple byte-level tokens, some subsequent tokens are determined almost uniquely, leading to higher probabilities than for English tokens.
Due to this characteristic of non-English languages, NegativeProb\textsubscript{pt} and other methods struggle to identify them as novelties.
In contrast, the contrastive score focuses on the difference in the log probability rather than their absolute values, performing robustly on both types of novelties.

Table \ref{tbl:extraction} (bottom) presents the results for the fine-tuning datasets of Falcon-RW. 
Similar to the results of OpenLLaMA, the contrastive score effectively distinguishes novelties from in-distribution examples across both datasets. 
Even when the in-distribution examples are not directly sourced from the pre-training dataset, the contrastive score consistently performs well in detecting novel examples.

\subsection{Generation of Novel Examples}
\label{sec:gen}

In this section, we assess our method, CGE, in the task of \emph {novelty discovery through generation}, where we aim to identify novel properties of a fine-tuning dataset by generating examples that illustrate these properties.
We demonstrate that CGE can discover a wide variety of novel characteristics that are hardly detected by simply sampling from the fine-tuned model.

\paragraph{Experimental Setup}

In our experiment, we generated 100 texts by each method and evaluated them using two metrics: detection and coverage rate.
Detection rate represents the percentage of generated texts that are identified as novel examples. A higher detection rate indicates that the method is more effective at generating novelties.
Coverage rate measures how well the generated texts cover novel examples across different domains. 
As previously explained, non-English languages, toxic texts, and source code are each categorized into 10 distinct domains. 
The coverage rate reflects the number of different domains that are represented in the generated texts.

To assess the content of the generated texts, we used the instruction-tuned LLaMA 3 (70B) model \citep{dubey2024llama}. 
We evaluated whether the texts were toxic, non-English, or programming languages, and further classified them into appropriate domains.
The prompts used for the evaluation are shown in Appendix \ref{app:experiment}.
Using the fine-tuning dataset, we evaluated the classification performance of LLaMA 3. 
The model was able to detect toxic text with 99.1\% accuracy and classify the target group of toxic text with 95.5\% accuracy. 
For non-English text, LLaMA 3 achieved 100\% accuracy in detecting and classifying the languages.
In the case of source code, it was able to detect code with 96.4\% accuracy and classify the programming languages with the same accuracy.

Since a validation set with ground-truth novel examples is typically unavailable, hyperparameter tuning for each experiment is impractical. 
Therefore, we set the hyperparameters based on the experimental results obtained from non-English text in OpenLLaMA, and applied the same values acrosss all subsequent experiments. 
Specifically, we used a plausibility constraint with $\alpha \!=\! 0.01$ and beam sampling with a beam size of four.
Appendix \ref{app:hyperparameter} shows that our results remain consistent across different hyperparameters.

\begin{table*}[t]
\caption{Performance on generating novel examples from fine-tuned models. The average and standard deviation across four runs are reported. The highest values are highlighted in \textbf{bold}.}
\label{tbl:generation}
\begin{center}
\begin{tabular}{llllll}
\toprule
\multirow{2}{*}{Model: OpenLLaMA}&\multicolumn{2}{c}{Non-English text}& &\multicolumn{2}{c}{Toxic text} \\ 
\cmidrule{2-3} \cmidrule{5-6}
 & Detection & Coverage &  & Detection & Coverage \\
\midrule
Sampling from fine-tuned models & 0.28±0.03 & 0.65±0.05 &  & 0.33±0.05 & \textbf{0.93±0.08} \\
CGE (static) & \textbf{0.99±0.01} & 0.55±0.05 &  & \textbf{0.78±0.06} & 0.75±0.05 \\
CGE (iterative) & 0.18±0.03 & \textbf{0.82±0.11} &  & 0.35±0.03 & 0.83±0.08 \\
\bottomrule
\end{tabular}
\end{center}
\begin{center}
\begin{tabular}{llllll}
\toprule
\multirow{2}{*}{Model: Falcon-RW}&\multicolumn{2}{c}{Non-English text}& &\multicolumn{2}{c}{Source code} \\ 
\cmidrule{2-3} \cmidrule{5-6}
 & Detection & Coverage &  & Detection & Coverage \\
\midrule
Sampling from fine-tuned models & 0.01±0.00 & 0.12±0.04 &  & 0.06±0.03 & 0.25±0.09 \\
CGE (static) & \textbf{0.53±0.05} & 0.43±0.04 &  & \textbf{0.92±0.02} & 0.62±0.04 \\
CGE (iterative) & 0.14±0.08 & \textbf{0.55±0.11} &  & 0.42±0.17 & \textbf{0.90±0.07} \\
\bottomrule
\end{tabular}
\end{center}
\end{table*}

\begin{figure}[t!]
\centering
\vspace{-\baselineskip}
\begin{minipage}{0.495\linewidth}
\centering
\includegraphics[width=0.9\linewidth]{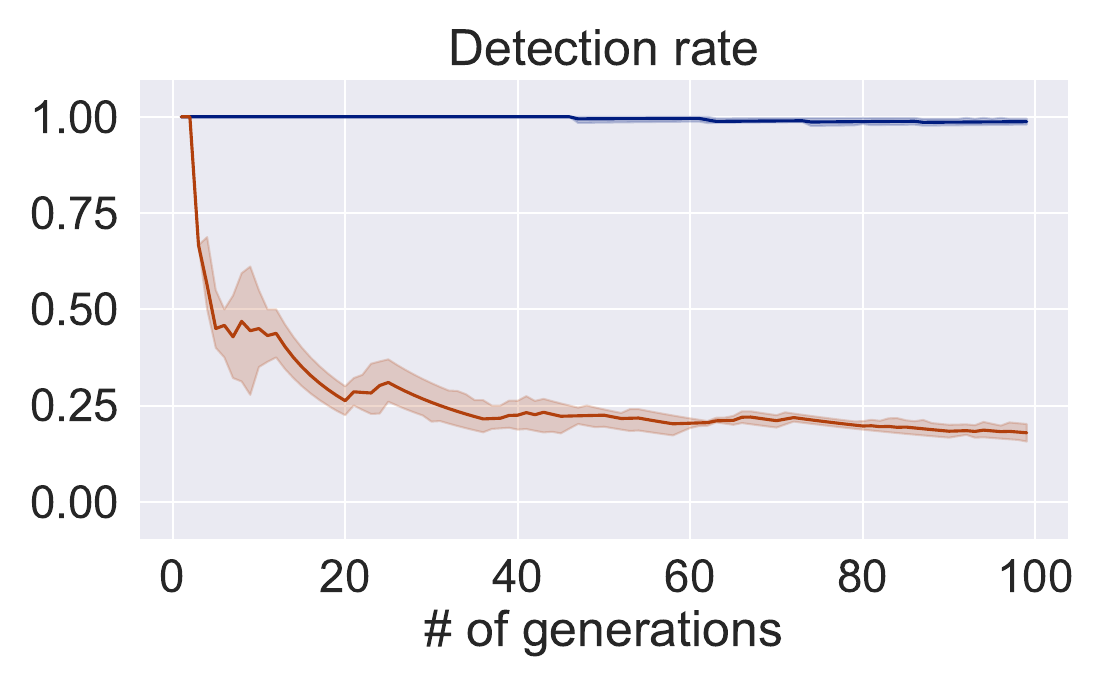}
\end{minipage}
\begin{minipage}{0.495\linewidth}
\centering
\includegraphics[width=0.9\linewidth]{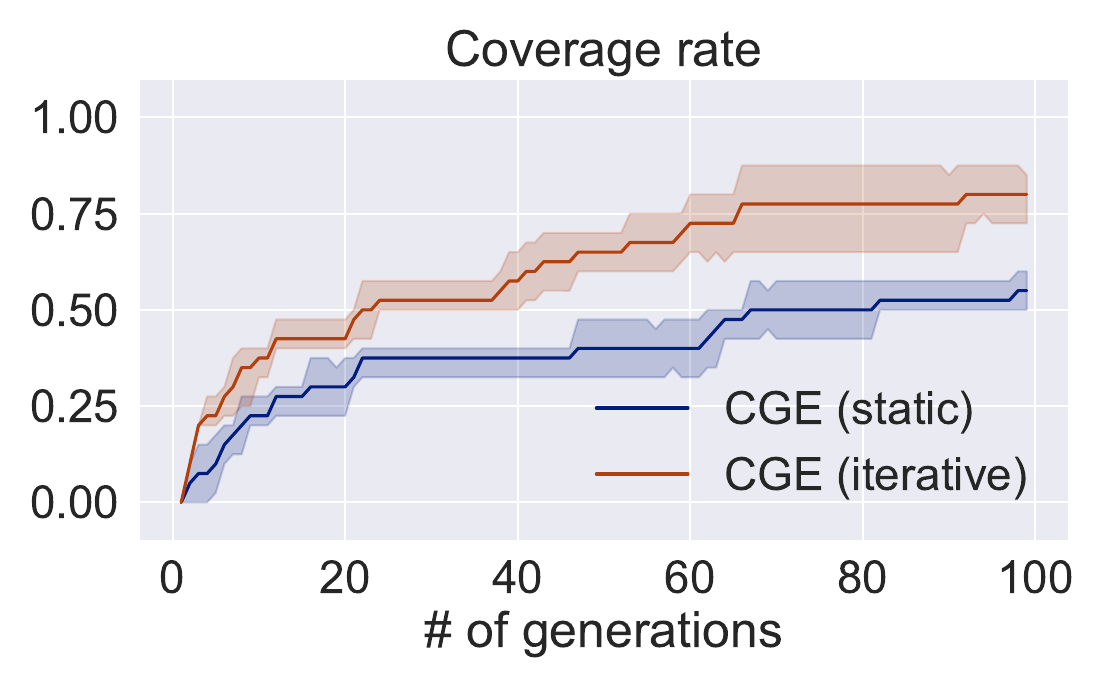}
\end{minipage}
\caption{Change in the detection and coverage rate across the different number of generated examples for the non-English dataset of OpenLLaMA. The line represents the average across four runs, and the shaded area corresponds to 95\% confidence region.}
\label{fig:change}
\end{figure}

\paragraph{Results}

We present the performance of generating novel examples from fine-tuned OpenLLaMA and Falcon-RW in Table \ref{tbl:generation}.
Sampling was conducted four times with different random seeds, and the average and standard deviation across these runs are reported.

When sampling directly from the fine-tuned models, we observed a low proportion of novel examples, resulting in a considerably lower detection rate. 
In contrast, CGE significantly improved both the detection and coverage rates, although a trade-off between the two metrics was apparent.
The static version achieved a notably higher detection rate, surpassing 90\% for non-English text as for OpenLLaMA and source code in Falcon-RW's fine-tuning dataset. 
However, the coverage rate was relatively low, around 60\%, indicating that fewer novel domains were being captured.
The iterative version substantially improved the coverage rate, exceeding 90\% for source code and over 80\% for non-English and toxic text in OpenLLaMA. 
However, this increase in coverage came at the cost of the detection rate. 
As the iterative version prevents the generation of previously seen examples, it allows the model to generate more in-distribution examples instead, which results in a lower detection rate.
In practical terms, this means that with the iterative version, the analyst will lose some time reviewing non-novel examples but will uncover a broader range of novel phenomena.

Figure \ref{fig:change} illustrates the change in the detection and coverage rate for varying numbers of generated examples in the non-English dataset of OpenLLaMA.
For the iterative version, the detection rate starts relatively high but steadily drops, whileit remains stable at almost 100\% for the static version.
In contrast, the coverage rate increases substantially for the iterative version, enhancing the diversity of generated novel examples.
This trade-off between the quantity and diversity of discovered novelties underscores the difficulty of the task.

\begin{figure}[t!]
\centering
\begin{minipage}{0.495\linewidth}
\centering
\includegraphics[width=.9\linewidth]{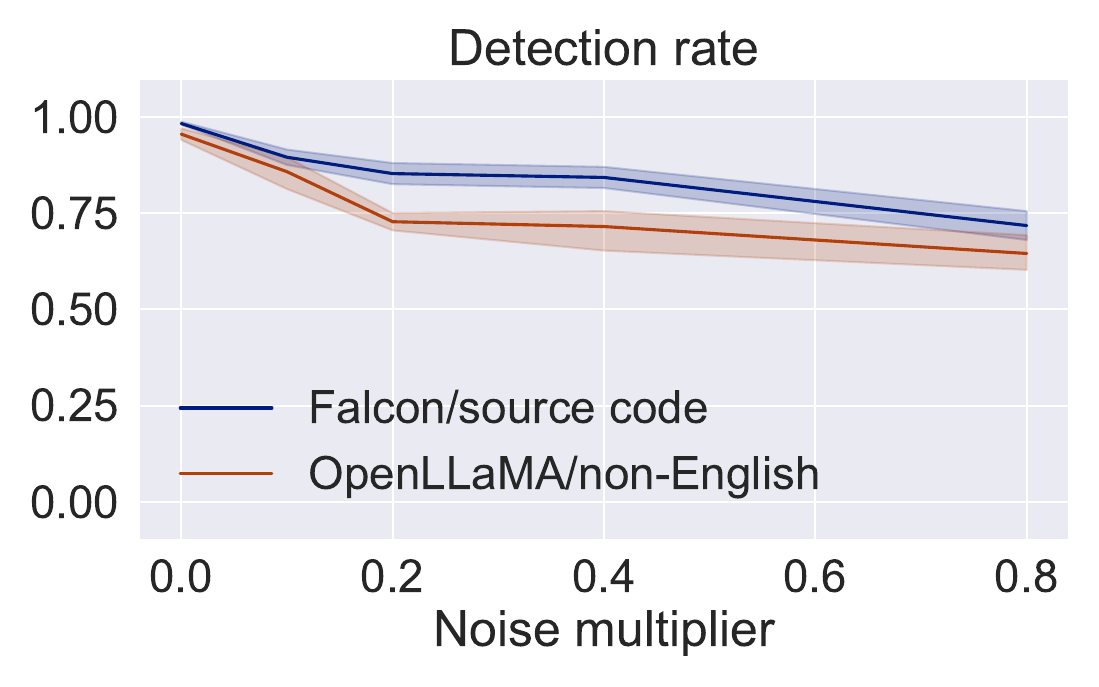}
\end{minipage}
\begin{minipage}{0.495\linewidth}
\centering
\includegraphics[width=.9\linewidth]{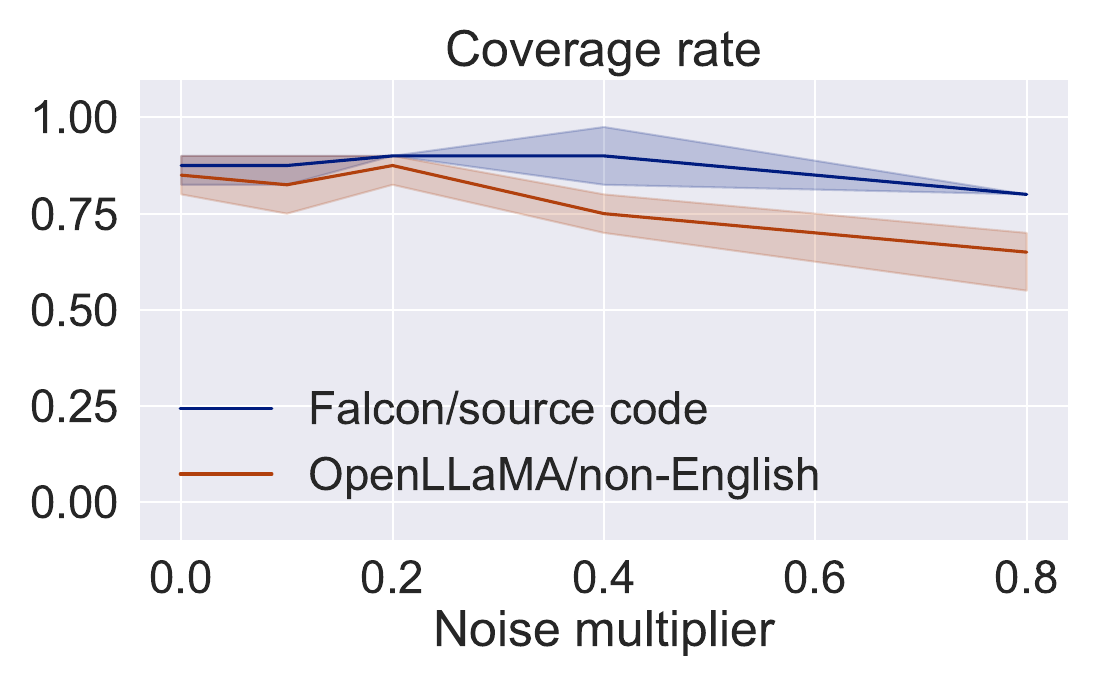}
\end{minipage}

\caption{Change in the detection and coverage rate across different values of noise multiplier. The line denotes the average across four runs, and the shaded area corresponds to 95\% confidence region.}
\label{fig:dp}
\end{figure}

\subsection{Effectiveness for Differentially Private Fine-tuned Models}

In this section, we demonstrate that CGE is also effective for models fine-tuned with differential privacy (DP) techniques. 
DP techniques are frequently used to protect sensitive data from privacy attacks, such as training data reconstruction or membership inference. 
Moreover, in practical deployments of DP, model designers and data analysis often lack access to the underlying data, rendering standard data analysis techniques infeasible \citep{garrido2023lessons, sarathy2023don}. 
While DP training reduces memorization, which poses additional challenges for gradient-based concept exploration (CGE), we demonstrate that CGE can still uncover novel features from fine-tuned models.

\paragraph{Experimental Setup}

We employ DP-Adam, a variant of DP-SGD \citep{song2013stochastic, bassily2014private, abadi2016deep}, which is widely used for DP fine-tuning and has been applied to language models in prior studies \citep{yu2022differentially, li2022large}. 
DP-Adam perturbs the gradients of training examples by clipping the per-example gradient norm and adding Gaussian noise, reducing the influence of individual training examples on the fine-tuned model. 
We fine-tuned a pre-trained model using DP-Adam, adjusting the strength of the Gaussian noise by setting different noise multipliers. 
We then assessed how the detection and coverage rates change as the noise multiplier increases.

As \citet{yu2022differentially} demonstrated, parameter-efficient fine-tuning methods, such as Low-Rank Adaptation  \citep[LoRA;][]{hu2022lora}, are more effective than updating all model parameters during DP fine-tuning. 
Following this study, by combining DP-Adam with LoRA, we fine-tuned OpenLLaMA on the RedPajama dataset augmented with non-English texts, and Falcon-RW on the English Wikipedia dataset augmented with source code. 
We injected trainable LoRA matrices into key, query, value, and linear transformation layers in the self-attention block. 
The intermediate representation dimension is set to $r = 8$ with a scaling factor of $\alpha = 16$, and the model is fine-tuned for three epochs with a learning rate of 5e-4.
For DP, we set the privacy budget $\delta$ to $1/n$, where $n$ is the size of the fine-tuning dataset, and adjusted the noise multiplier to 0.0, 0.1, 0.2, 0.4, and 0.8.

\paragraph{Results}

Figure \ref{fig:dp} presents the change in detection and coverage rate across different noise multipliers.
We generated 100 texts using CGE and evaluated the generated texts as described in Section \ref{sec:gen}.
Introducing DP led to a decline in the detection rate, though the impact was not substantial even with higher noise multipliers. 
Similarly, DP had a marginal impact on the coverage rate, which remained above 60\% for both models. 
These findings suggest that our methods can reliably uncover novel examples even when models are fine-tuned with DP techniques.

\section{Related Work}

\paragraph{Dataset Exploration}
Exploring the properties of datasets is a crucial step in model development. 
Prior work has mainly focused on providing methods and tools to directly inspect datasets.

 detection techniques use trained models to identify novel examples that deviate from training data distribution \citep{lee2018simple, yang2024generalized}. 
For instance, the maximum softmax score \citep{hendrycks2017baseline} and its extension \citep{liang2018enhancing, hsu2020generalized} detect novel examples by identifying low-confidence predictions. 
Likewise, \citet{liu2020energy, huang2021importance} leverage energy functions or gradient norms to detect novelties effectively.

Another research direction focuses on improving dataset transparency. 
\citet{piktus-etal-2023-roots, piktus-etal-2023-gaia, elazar2024whats} offer tools to inspect large text corpora, enabling users to identify potential data contamination or biases by directly accessing and querying the training data. 
Similarly, \citet{marone2023data, zhou2024oasis} have developed fast, space-efficient querying systems and customizable rule-based methods for filtering and optimizing training data.

Our work addresses real-world scenarios where fine-tuning is conducted on massive, noisy, and confidential datasets, making direct inspection impractical. 
We focus on problems where we aim to infer dataset properties by analyzing a model’s behavior without direct access to the data. 
Recent works \citep{shi2024detecting, golchin2024time} have introduced a similar task, where they detect data contamination by examining a model’s outputs without dataset access. 
Aligning with these works, we introduced a novel task, which aims to identify novel examples in a fine-tuning dataset that deviates from the pre-training data distribution without dataset access.

\paragraph{Contrastive Decoding}
Contrastive decoding is a method for generating text that highlights differences between the predictions of two models: an expert model (e.g., a large model or non-toxic model) and an amateur model (e.g., a small model or toxic model). 
The objective is to generate text favored by the expert model while simultaneously discouraging the preferences of the amateur model. 
The utility of contrastive decoding and its variants have been demonstrated in various applications, such as ensuring the safety of the generated text \citep{liu-etal-2021-dexperts, xu-etal-2024-safedecoding, shi-etal-2024-navigating, zhong-etal-2024-rose}, improving the quality of generation \citep{li-etal-2023-contrastive, o2023contrastive}, or instruction tuning \citep{liu2024tuning, gao2024linear}.

This work extends the use of contrastive decoding to explore novel features within fine-tuning datasets.
By contrasting the fine-tuned model against the pre-trained model, our method identifies sequences that illustrates novelties in the fine-tuning data. 
We also introduced an iterative version that could be beneficial in other scenarios where contrastive decoding is applied.

\paragraph{Dataset Distillation}
Dataset distillation is a technique aimed at creating a small, representative synthetic dataset that retains the core properties of a much larger dataset. 
While most methods were developed for image classification tasks, recent efforts have explored their application in text classification. 
\citet{li2021data, sucholutsky2021soft, maekawa-etal-2023-dataset, maekawa-etal-2024-dilm} have extended dataset distillation to text classification tasks, despite the complexity of dealing with discrete sequence data. 
However, these methods often face challenges, such as the cost of calculating second-order derivatives, making them less scalable for larger models.
Furthermore, these works only consider text classification datasets and have difficulty being used for language modeling datasets.

CGE is closely related to dataset distillation, but shifts focus toward discovering novelties. 
With the first-order approximation, CGE can be reduced to a form of dataset distillation, but with significantly lower computational cost.
Our method can be applicable to language modeling datasets, and the distilled dataset consists of interpretable text.
It also has the potential to serve as a dataset compression technique, aiming to create a smaller training corpus that resembles a large-scale corpus.

\section{Conclusion}

In this paper, we introduced the task of \emph{novelty discovery through generation}, which aims to identify novel properties in a fine-tuning dataset without having direct access to the data. 
As a simple solution to this task, we proposed \emph{Contrastive Generative Exploration} (CGE), which uncovers novel properties in fine-tuning datasets by generating examples that represent these properties. 
Our experimental results demonstrated that CGE effectively detects novel properties in both extraction and generation settings. 
Additionally, we showed that our method is robust to the noise introduced by DP techniques when models are fine-tuned using DP-Adam, proving its efficacy even in scenarios where access to the data is restricted. 
However, we also indicated that there exists a trade-off between the quantity and diversity of the discovered novelties, underscoring the inherent challenge of the task.
In future work, we anticipate the development of methods that can more effectively resolve this trade-off. 
Moreover, we look forward to experiments conducted using real-world datasets, to drive the development of more practical and robust approaches.

\subsubsection*{Acknowledgments}

MI is partially supported by JST CREST JPMJCR21D1, NEDO JPNP20006, and JSPS KAKENHI 23K16940, Japan.
IT is supported by the Dutch National Science Foundation (NWO Vici VI.C.212.053).

\bibliography{main, anthology_0, anthology_1}
\bibliographystyle{iclr2025_conference}

\appendix
\newpage
\section{Appendix}

\begin{table*}[t]
\caption{Prompts used for LLaMA 3 to evaluate generated text.}
\label{tbl:prompt}
\begin{center}
\begin{tabularx}{\textwidth}{p{1.8cm}L}
\toprule
Dataset & Prompt \\
\midrule
Toxic text & Determine the target of the hate speech. Select the most representative category from the following categories: asian, muslim, physical\_disability, jewish, native\_american, lgbtq, women, mental\_disability, latino, black. Answer `no' if the text is not hate speech. Answer with a single category only and do not generate anything else. \\ [55pt]
Non-English & Select the language of the text from the following choices: ja, zh, fa, ar, he, tr, id, ko, vi, th, en. Answer `en' if the language does not exist in the choices. Answer with a single choice only and do not generate anything else. \\[30pt]
Source code & Select the programming language of the text from the following choices: JavaScript, Java, C, Python, Ruby, TypeScript, Shell, GO, SQL, Perl. Answer `no' if the text does not correspond to any programming language. Answer with a single choice only and do not generate anything else. \\
\bottomrule
\end{tabularx}
\end{center}
\end{table*}

\subsection{Baseline Methods for Extraction Setting}
\label{app:baseline}

The following methods are used as the baseline methods for the experiments in the extraction setting.
Higher scores indicate an example is more likely to be novel, while lower scores suggest the example is in-distribution.

\paragraph{MSP \citep{hendrycks2017baseline}} Maximum softmax probability for each token $t$ in an example: $-\sum_{t} \max_x \log p(x_t\!=\!x | x^{<t}; \bm{\theta}_\mathrm{pt})$.

\paragraph{Energy \citep{liu2020energy}} Energy score (the denominator of the softmax activation) for each prediction: $\sum_{t} \log \sum_x \exp(f(x| x^{<t};\bm{\theta}_\mathrm{pt}))$ where $f(x|x^{<t};\bm{\theta}_\mathrm{pt})$ is the logit of $x$ for the $t$-th token.

\paragraph{GradNorm \citep{huang2021importance} } The norm of gradient where the label for each prediction is uniformly distributed: $\|\nabla_{\bm{\theta}} \sum_{t} D_{\mathrm{KL}}[\bm{u}||p(\cdot| x^{<t}; \bm{\theta}_\mathrm{pt})]\|$ where $\bm{u}$ is the uniform distribution over tokens.

\paragraph{NegativeProb\textsubscript{pt}} Negative log-probability computed by the pre-trained model: $-\log p(\bm{x}; \bm{\theta}_\mathrm{pt})$.

\paragraph{Prob\textsubscript{ft}} Log-probability of tokens computed by the fine-tuned model: $\log p(\bm{x}; \bm{\theta}_\mathrm{ft})$. 

\paragraph{GradientNorm\textsubscript{pt}} The norm of gradient w.r.t. the pre-trained model: $\|\nabla_{\bm{\theta}} \log p(\bm{x}; \bm{\theta}_\mathrm{pt})\|$.

\subsection{Experimental Details}
\label{app:experiment}
Table \ref{tbl:prompt} shows the prompts used for the evaluation of generated texts by using LLaMA 3.\footnote{\url{https://huggingface.co/meta-llama/Meta-Llama-3-70B-Instruct}}
Given the promt and the generated text, the probability of each domain is computed. The domain with the highest probability is selected as an answer.

\begin{table*}[t]
\caption{Performance of CGE on discovering novelties when varying the hyperparameters. The average and standard deviation across four runs are reported.}
\label{tbl:hp}
\begin{center}
\begin{tabular}{llllll}
\toprule
Model: OpenLLaMA&\multicolumn{2}{c}{Static version}& &\multicolumn{2}{c}{Iterative version} \\ 
\cmidrule{2-3} \cmidrule{5-6}
Dataset: Non-English Text & Detection & Coverage &  & Detection & Coverage \\
\midrule
beam sampling, alpha=0.01 & 0.99±0.01 & 0.55±0.05 &  & 0.18±0.03 & 0.82±0.11 \\
sampling, alpha=0.01 & 0.95±0.02 & 0.70±0.07 &  & 0.22±0.01 & 0.93±0.08 \\
beam sampling, alpha=0.1 & 1.00±0.00 & 0.25±0.05 &  & 0.26±0.03 & 0.62±0.04 \\
\bottomrule
\end{tabular}
\end{center}
\end{table*}

\subsection{Hyperparameter Sensitivity}
\label{app:hyperparameter}

Table \ref{tbl:hp} shows the performance of CGE on discovering novel examples when varying the hyperparameters.
The trend does not change significantly with different hyperparameters. 
The static version consistently achieves a high detection rate, while the iterative version improves the coverage rate.
With a large alpha, the diversity of generated text decreases due to the adaptive plausibility constraint, resulting in a relatively lower coverage rate.

\end{document}